\documentclass{article}

\usepackage[final,nonatbib]{nips_2017}

\usepackage[utf8]{inputenc} 
\usepackage[T1]{fontenc}    
\usepackage{hyperref}       
\usepackage{url}            
\usepackage{booktabs}       
\usepackage{amsfonts}       
\usepackage{nicefrac}       
\usepackage{microtype}      
\usepackage{subfiles}
\usepackage{subcaption}
\usepackage{lipsum}
\usepackage{multirow}
\usepackage{color}
\usepackage{graphicx}
\usepackage{amsmath}
\usepackage{dsfont}
\usepackage{cite}
\definecolor{orangewecanread}{rgb}{1.0, 0.6, 0.4}

\title{``Found in Translation'': Predicting Outcomes of Complex Organic Chemistry Reactions using Neural Sequence-to-Sequence Models}

\newcommand*\samethanks[1][\value{footnote}]{\footnotemark[#1]}
\author{
	Philippe Schwaller\thanks{Equal contributors}, \hspace{2mm} Théophile Gaudin\samethanks, \hspace{2mm} Dávid Lányi, \hspace{2mm} Costas Bekas, \hspace{2mm} Teodoro Laino \\ 
  IBM Research, Zurich\\
  \texttt{\{phs,tga,dla,bek,teo\}@zurich.ibm.com} \\
}

\begin{document}

\maketitle

\begin{abstract}
	There is an intuitive analogy of an organic chemist's understanding of a compound and a language speaker's understanding of a word. Consequently, it is possible to introduce the basic concepts and analyze potential impacts of linguistic analysis to the world of organic chemistry. In this work, we cast the reaction prediction task as a translation problem by introducing a template-free sequence-to-sequence model, trained end-to-end and fully data-driven. We propose a novel way of tokenization, which is arbitrarily extensible with reaction information. With this approach, we demonstrate results superior to the state-of-the-art solution by a significant margin on the top-1 accuracy. Specifically, our approach achieves an accuracy of 80.3\% without relying on auxiliary knowledge such as reaction templates. Also, 65.4\% accuracy is reached on a larger and noisier dataset.


\end{abstract}

\section{Introduction}
After nearly 200 years of documented research, the synthesis of organic molecules remains one of the most important tasks in organic chemistry. The construction of a target molecule from a set of existing reactants and reagents via chemical reactions is attracting much attention because of its economical implications. 

Multiple efforts have been made in the past 50 years to rationalize the large number of chemical compounds and reactions identified, which form the large knowledge bases for solving synthetic problems. In 1969, Corey and Wipke \cite{Corey1969} demonstrated that both synthesis and retrosynthesis could be performed by a machine. Their pioneering contribution involved the use of handcrafted rules made by experts, which are commonly known as reaction templates. The templates encode the local changes to the atoms' connectivity under certain conditions accounting for various subtleties of retrosynthesis. A similar algorithm emerged in the late 1970s \cite{Salatin1978} which also requires a set of expert rules. Unfortunately, rules writing is a tedious task, both time and labor-intensive, and may not cover the entire domain for complex organic chemistry problems. In such cases, profound chemical expertise is still required, and the solutions are usually developed by trained organic chemists. However, it can be extremely challenging even for them to synthesize a relatively complex molecule, which may take several reaction steps to construct. In fact, navigating the chemical space of drug-related compounds by relying only on intuition may turn a synthesis into a nearly impossible task, especially if the problem is slightly outside the expert's knowledge.

Other approaches extract reaction templates directly from data \cite{Satoh1995, Satoh1996, Segler2017, Coley2017}. In this specific context, candidate products are generated from the templates and then are ranked according to their likelihood. Satoh and Funatsu \cite{Satoh1995, Satoh1996} used various hard-coded criterion to perform the ranking whereas more recent approaches \cite{Segler2017, Coley2017} used a deep neural network. However, these types of approaches are fundamentally dependent on the rule-based system component and thus inherit some of its major limitations. In particular, these approaches do not produce sufficiently accurate predictions outside of the training domain. 

Nevertheless, the class of algorithms \cite{Corey1969, Salatin1978, Satoh1995, Satoh1996, Segler2017, Coley2017} that is based on rules manually encoded by human experts or automatically derived from a reaction database is not the only way to approach the problem of organic synthesis. A second approach to predicting reactions has been to leverage the advancements in computational chemistry to predict energy barriers of a reaction based on first-principle calculations \cite{Dolbier1996,Mondal2013}. Although it is possible to reach very accurate levels of predictions for small systems, it is still a very computationally intensive task. Therefore, it is limited to applications of purely academic interest.

One way to view the reaction prediction task is to cast it as a translation problem, where the objective is to map a text sequence that represents the reactants to a text sequence representing the product. Molecules can equivalently be expressed as text sequences in line notation format, such as the simplified molecular-input line-entry system (SMILES) \cite{Weininger1988}. Intuitively, there is an analogy between a chemist's understanding of a compound and a language speaker's understanding of a word. No matter how imaginative such an analogy is, it was only very recently that a formal verification was proved \cite{Cadeddu2014}. Cadeddu et al. \cite{Cadeddu2014} showed that organic molecules contain fragments whose rank distribution is essentially identical to that of sentence fragments. The immediate consequence of this discovery is that the vocabulary of organic chemistry and human language follow very similar laws, thus introducing the basic concepts and potential impact of linguistics-based analyses to a general chemical audience. It has already been shown that a text representation of molecules has been effective in chemoinformatics \cite{Gomez-Bombarelli2016,Jastrzebski2016,Kusner2017,Bjerrum2017,Segler2017a}. This has strengthened our belief that the methods of computational linguistics can have an immense impact on the analysis of organic molecules and reactions.
 
In this work, we build on the idea of relating organic chemistry to a language and explore the application of state-of-the-art neural machine translation methods, which are sequence-to-sequence (seq2seq) models. A similar approach was recently suggested, but its application was limited to textbook reactions \cite{Nam2016}. Here, we intend to solve the forward-reaction prediction problem, where the starting materials are known and the interest is in generating the products. We propose a novel way of tokenization that is arbitrarily extensible with reaction information. The overall network architecture is simple, and the model is trained end-to-end, fully data-driven and without additional external information. With this approach, we outperform current solutions using their own training and test sets \cite{Jin2017} by achieving a top-1 accuracy of 80.3\% and set a first score of 65.4\% on a noisy single product reactions dataset extracted from US patents.

\section{Related Work}
\subsection{Template-based reaction prediction}
Template-based reaction prediction methods have been widely researched in the past couple of years \cite{Wei2016,Segler2017,Coley2017}. Wei et al. \cite {Wei2016} used a graph-convolution neural network proposed by Duvenaud et al. \cite{Duvenaud2015} to infer fingerprints of the reactants and reagents. They trained a network on the fingerprints to predict which reaction templates to apply to the reactants. Segler and Waller \cite{Segler2017} built a knowledge graph using reaction templates and discovered novel reactions by searching for missing nodes in the graph. Coley et al. \cite{Coley2017} generated for a given set of reactants all possible product candidates from a set of reaction templates extracted from US patents \cite{Lowe2012} and predicted the outcome of the reaction by ranking the candidates with a neural network. One major advancement by Segler and Waller \cite{Segler2017} and Coley et al. \cite{Coley2017} was to consider alternative products as negative examples. Recently, Segler and Waller \cite{Segler2017b} introduced a neural-symbolic approach. They extracted reaction rules from the commercially available Reaxys database. Then, they trained a neural network on molecular fingerprints to prioritize rules and combined the network with a Monte Carlo tree search to overcome the scalability issues of other template-based methods. In any case, template-based methods have the limitation that they cannot predict anything outside the space covered by the previously extracted templates. 

\subsection{Template-free reaction prediction}
A first template-free approach was introduced by Kayala et al. \cite{Kayala2012}. Using fingerprints and hand-crafted features, they predicted a series of mechanistic steps to obtain one reaction outcome. Owing to the sparsity of data on such mechanistic reaction steps, the dataset was self-generated with a template-based expert system. Recently, Jin et al. \cite{Jin2017} used a novel approach based on Weisfeiler--Lehman Networks (WLN). They trained two independent networks on a set of 400,000 reactions extracted from US patents. The first WLN scored the reactivity between atom pairs and predicted the reaction center. All possible bond configuration changes were enumerated to generate product candidates. The candidates that were not removed by hard-coded valence and connectivity rules are then ranked by a Weisfeiler--Lehman Difference Network (WLDN). Their method achieved a top-1 accuracy of 74.0\% on a test set of 40,000 reactions. Jin et al. \cite{Jin2017} claimed to outperform template-based approaches by a margin of 10\% after augmenting the model with the unknown products of the initial prediction to have a product coverage of 100\% on the test set. Although the code is not yet public, the dataset with the exact training, validation and test split have been released\footnote{\url{https://github.com/wengong-jin/nips17-rexgen}}. The complexity of the reaction prediction problem was significantly reduced by removing the stereochemical information. 

\subsection{Seq2seq models in organic reaction prediction and retrosynthesis}

The closest work to ours is that of Nam and Kim \cite{Nam2016}, who also used a template-free seq2seq model to predict reaction outcomes. Whereas their network was trained end-to-end on patent data and self-generated reaction examples, they limited their predictions to textbook reactions. Their model was based on the Tensorflow translate model (v0.10.0) \cite{Abadi2016}, from which they took the default values for most of the hyperparameters. 

Retrosynthesis is the opposite of reaction prediction. Given a product molecule, the goal is to find possible reactants. This is a considerably more difficult task for a seq2seq model and was approached by Liu et al. \cite{Liu2017}. They used a set of 50,000 reactions extracted and curated by Schneider et al. \cite{Schneider2016}. Although the stereochemical information was included, the reactions were classified, which means that the dataset contained only common reaction types. Overall, none of the previous works was able to demonstrate the superiority of seq2seq models.


What we observe in general is that there is always a tradeoff between coverage and accuracy. In fact, whenever reactions that do not work well with the model are removed under the assumption that they are erroneous, the model's accuracy will improve. This calls for open datasets. The only fair way to compare models is to use datasets to which identical filtering was applied or where the reactions that the model is unable to predict are counted as false predictions.

\section{Dataset}
All the openly available chemical reaction datasets were derived in some form from the patent text-mining work of Daniel M. Lowe \cite{Lowe2012}. Lowe's dataset has recently been updated and contains data extracted from US patents grants and applications dating from 1976 to September 2016 \cite{Lowe2017}. What makes the dataset particularly interesting is that the quality and noise correspond well to the data a chemical company might own. The portion of granted patents is made of 1,808,938 reactions, which are described using SMILES \cite{Weininger1988}. 

Looking at the original patent data, it is surprising that a complex chemical synthesis process consisting of multiple steps, performed over hours or days, can be summarized in a simple string. Such reaction strings are composed of three groups of molecules: the reactants, the reagents, and the products, which are separated by a `$>$' sign. The process actions and reaction conditions, for example, have been neglected so far. 

To date, there is no standard way of filtering duplicates, incomplete or erroneous reactions in Lowe's dataset. We kept the filtering to a minimum to show that our network is able to handle noisy data. We removed 720,768 duplicates by comparing reaction strings without atom mapping and an additional 780 reactions, because the SMILES string could not be canonicalized with RDKit \cite{Landrum2017}, as the explicit number of valence electrons for one of the atoms was greater than permitted. We took only single product reactions, corresponding to 92\% of the dataset, to have distinct prediction targets. Although this is a current limitation in the training procedure of our model, it could be easily overcome in the future, for example by defining a specific order for the product molecules. Finally, the dataset was randomly split into training, validation and test sets (18:1:1)\footnote{\url{https://ibm.box.com/v/ReactionSeq2SeqDataset}}. Reactions with the same reactants, but different reagents and products were kept in the same set.  
\begin{table}[!ht]
\centering
\caption{Overview of the datasets used in this work. Jin's is derived from Lowe's grants dataset.}
    \begin{tabular}{l | rrr | r }
        \toprule 
        Reactions in & train & valid & test & total \\
        \midrule
        Lowe's grants set \cite{Lowe2017} &&&& 1,808,938 \\
        \ \ \ without duplicates &&&& 1,088,170 \\
        \ \ \  with single product&902,581 & 50,131 & 50,258 & 1,002,970 \\
         \midrule
        Jin's USPTO set \cite{Jin2017} & 409,035 & 30,000 & 40,000& 479,035  \\
         \ \ \ with single product &  395,496 & 29,075 & 38,647& 463,218 \\
         \bottomrule
    \end{tabular}
    \label{tab:dataset}
\end{table}

To compare our model and results with the current state of the art, we used the USPTO set recently published by Jin et al. \cite{Jin2017}. It was extracted from Lowe's grants dataset \cite{Lowe2017} and contains 479,035 atom-mapped reactions without stereochemical information. We restricted ourselves to single product reactions, corresponding to 97\% of the reactions in Jin's USPTO set. An overview of the datasets taken as ground truths for this work is shown in Table \ref{tab:dataset}. 

\begin{table}[!ht]
    \caption{Data preparation steps to obtain source and target sequences. The tokens are separated by a space and individual molecules by a point token.}
    \begin{tabular}{ll|l}
        \toprule
        &Step & Example (entry 23738, Jin's USPTO test set \cite{Jin2017}) \\
        &&reactants $>$ reagents $>$ products \\
        \midrule
         1) &Original string &  [Cl:1][c:2]1[cH:3][c:4]([CH3:8])[n:5][n:6]1[CH3:7].[OH:14][N+:15] \\
          && ([O-:16])=[O:17].[S:9](=[O:10])(=[O:11])([OH:12])[OH:13]$>>$[Cl:1] \\
          && [c:2]1[c:3]([N+:15](=[O:14])[O-:16])[c:4]([CH3:8])[n:5][n:6]1[CH3:7] \\ 
        \midrule
        2)& Reactants and & [Cl:1][c:2]1[cH:3][c:4]([CH3:8])[n:5][n:6]1[CH3:7].[OH:14][N+:15] \\
         &reagent & ([O-:16])=[O:17]$>$[S:9](=[O:10])(=[O:11])([OH:12])[OH:13]$>$[Cl:1] \\
         &separation& [c:2]1[c:3]([N+:15](=[O:14])[O-:16])[c:4]([CH3:8])[n:5][n:6]1[CH3:7] \\
         \midrule
        3) &Atom-mapping& Cc1cc(Cl)n(C)n1.O=[N+]([O-])O \\
         & removal and &$>$O=S(=O)(O)O$>$\\
         & canonicalization &  Cc1nn(C)c(Cl)c1[N+](=O)[O-]\\
        \midrule
        4) &Reactants  & C c 1 c c ( Cl ) n ( C ) n 1 . O = [N+] ( [O-] ) O\\ 
         & and product & $>$O=S(=O)(O)O$>$  \\   
         & tokenization & C c 1 n n ( C ) c ( Cl ) c 1 [N+] ( = O ) [O-]\\ 
        \midrule
        5) &Reagent   & C c 1 c c ( Cl ) n ( C ) n 1 . O = [N+] ( [O-] ) O\\ 
         & tokenization &  $>$ A\_O=S(=O)(O)O $>$  \\
         & &  C c 1 n n ( C ) c ( Cl ) c 1 [N+] ( = O ) [O-]\\ 
        \midrule
        \midrule
        &Source & C c 1 c c ( Cl ) n ( C ) n 1 . O = [N+] ( [O-] ) O \\
        && $>$ A\_O=S(=O)(O)O \\
        \midrule
        &Target & C c 1 n n ( C ) c ( Cl ) c 1 [N+] ( = O ) [O-] \\
        \midrule
        \bottomrule
    \end{tabular}
    \label{tab:preprocessing}
\end{table}

\subsection{Data preprocessing}

To prepare the reactions, we first used the atom mappings to separate reagents from reactants. Input molecules with atoms appearing in the product were classified as reactants and the others without atoms in the product as reagents. Then, we removed the hydrogen atoms and the atom mappings from the reaction string, and canonicalized the molecules. Afterwards, we tokenized reactants and products atom-wise using the following regular expression:
\vspace{-0.1cm}
\begin{quote}
\texttt{token\_regex = "(\textbackslash[[\textasciicircum \textbackslash]]+]|Br?|Cl?|N|O|S|P|F|I|b|c|n|o|s|p|\textbackslash(|\textbackslash)| \textbackslash.|=|\#|-|\textbackslash+|\textbackslash\textbackslash\textbackslash\textbackslash|\textbackslash/|:|\textasciitilde|@|\textbackslash?|>|\textbackslash*|\textbackslash\$|\textbackslash\%[0-9]\{2\}|[0-9])"}.

\end{quote}
\vspace{-0.1cm}
As reagent atoms are never mapped into product atoms, we employed a reagent-wise tokenization using a set of the 76 most common reagents, according to the analysis in \cite{Schneider2016}. Reagents belonging to this set were added as distinct tokens after the first `$>$' sign, ordered by occurrence. Other reagents, which were not in the set, were neglected and removed completely from the reaction string. The separate tokenization would allow us to extend the reaction information and add tokens for reaction conditions without changing the model architecture. The final source sequences were made up of tokenized ``reactants > common reagents'' and the target sequence of a tokenized ``product''. The tokens were separated by space characters. The preprocessing steps together with examples are summarized in Table \ref{tab:preprocessing}. The same preprocessing steps were applied to all datasets. 



%
%
%
%
%
%
%
%
%
%
%
%
%
%
%
%
%
%
%
%

\section{Model}
\label{sec:model}
To map the sequence of the reactants/reagents to the sequence of the products, we adapted an existing implementation \cite{Zhao2017} with minor modifications. Our model architecture consists of two distinct recurrent neural networks (RNN) working together: (1) an encoder that processes the input sequence and emits its context vector $C$, and (2) a decoder that uses this representation to output a probability over a prediction. For these two RNNs, we rely on specific variants of long short-term memory (LSTM) \cite{Hochreiter1997} because they are able to handle long-range relations in sequences. An LSTM consists of units that process the input data sequentially. Each unit at each time step $t$ processes an element of the input $x_t$ and the network’s previous hidden state $h_{t-1}$. The output and the hidden state transition is defined by
\begin{align}
	i_t &= \sigma\left(W_i \cdot x_t + U_i \cdot h_{t-1} + b_i\right) \\
	f_t &= \sigma\left(W_f \cdot x_t + U_f \cdot h_{t-1} + b_f\right) \\
	o_t &= \sigma\left(W_o \cdot x_t + U_o \cdot h_{t-1} + b_o\right) \\
	c_t &= f_t \otimes c_{t-1} + i_t \otimes \tanh\left( W_c \cdot x_t + U_c \cdot h_{t-1} + b_c\right) \\
	h_t &= o_t \otimes \tanh \left( c_{t-1} \right), 
\end{align}
where $i_t$, $f_t$ and $o_t$ are the input, forget, and output gates; $c$ is the cell state vector; $W$, $U$ and $b$ are model parameters learned during training; $\sigma$ is the sigmoid function and $\otimes$ is the entry-wise product. For the encoder, we used a bidirectional LSTM (BLSTM) \cite{Graves2005}. A BLSTM processes the input sequence in both directions, so they have context not only from the past but also from the future. They comprise two LSTMs: one that processes the sequence forward and the other backward, with their forward and backward hidden states $\overrightarrow{h}_t$ and $\overleftarrow{h}_{t}$ for each time step. The hidden states of a BLSTM are defined as
\begin{align}
	h_t &= \{\overrightarrow{h}_t; \overleftarrow{h}_{t}\}. 
\end{align}
Thus we can formalize our encoder as 
\begin{align}
	C = f \left( W_e \cdot x_t, h_{t-1} \right),
\end{align}
where $f$ is a multilayered BLSTM; $h_t \in \mathds{R}^n$ are the hidden states at time $t$; $x_t$ is an element of an input sequence $x = \{ x_0, \dots, x_T\}$, which is a one-hot encoding of our vocabulary; and $W_e$ are the learned embedding weights. Generally, $C$ is a simple concatenation of the encoder's hidden states:
\begin{equation}
	C = \{ h_1, \dots, h_T \}
\end{equation}
The second part of the model -- the decoder -- predicts the probability of observing a product $\hat{y} = \{\hat{y}_1, \dots, \hat{y}_M \}$:
\begin{equation}
	P \left( \hat{y} \right) = \prod^{M}_{i=0} p \left( \hat{y}_i | \{ \hat{y}_1, \dots, \hat{y}_{i-1} \} \right)
\end{equation}
and for a single token $\hat{y}_i$:
\begin{equation}
	p \left( \hat{y}_i | \{ \hat{y}_1, \dots, \hat{y}_{i-1} \}, c_i \right) = g\left( \hat{y}_{i-1}, s_i, c_i \right),
\end{equation}
where $g$ is a stack of LSTM, which outputs the probability $\hat{y}_t$ for a single token; $s_i$ are the decoder's hidden states; and $c_i$ is a different context vector for each target token $y_i$. Bahdanau et al. \cite{Bahdanau2015} and Luong et al. \cite{Luong2015} proposed attention mechanisms, \emph{i.e.}, different ways for computing the $c_i$ vector rather than taking the last hidden state of the encoder $h_t$. We performed experiments using both models and describe Luong's method, which yielded the best overall results.

\subsection{Luong's Attention Mechanism}
To compute the context vector, we first have to compute the attention weights $\alpha$: 
\begin{align}
	\alpha_{it} &= \frac{
		\exp\left(
			s_i^\top\cdot W_{\alpha}\cdot h_t
		\right)
	}{
	\sum^{T}_{t'=0}\exp\left(
		s_i^\top\cdot W_{\alpha}\cdot h_{t'}
	\right)
	} \\
	c_i &= \sum^{T}_{t = 0}\alpha_{it} \cdot h_t,
\end{align}
The attention vector is then defined by
\begin{align}
	a_i &= \tanh \left( W_a \cdot \{ c_i ; s_i \} \right).
\end{align}
Both $W_{\alpha}$ and $W_a$ are learned weights. Then $a$ can be used to compute the probability for a particular token:\begin{equation}
	p \left( y_i | \{ y_1, \dots, y_{i-1} \}, c_i \right) =  \textrm{softmax} \left(W_p \cdot a_i \right),
\end{equation}
where $W_p$ are also the learned projection weights.

\subsection{Training Details}
During training, all parameters of the network were trained jointly using a stochastic gradient descent. The loss function was a cross-entropy function, expressed as
\begin{equation}
	\label{eqn:jesuisunebelleequation}
	H\left(y, \hat{y}\right) = - \sum_i y_i \log \left( \hat{y}_i \right)
\end{equation}
for a particular training sequence. The loss was computed over an entire minibatch and then normalized. The weights were initialized using a random uniform distribution ranging from $-0.1$ to 0.1. Every 3 epochs, the learning rate was multiplied by a decay factor. The minibatch size was 128. Gradient clipping was applied when the norm of the gradient exceeded 5.0. The teacher forcing method \cite{Williams1989} was used during training.

\section{Architecture \& Hyperparameter Search}

\paragraph{}
Finding the best-performing set of hyperparameters for a deep neural network is not trivial. As mentioned in \autoref{sec:model}, our model has numerous parameters that can influence both its training and its architecture. Depending on those parameters, the performance of the model can vary notably. In order to select the best parameters efficiently, we build a framework around scikit-optimize \cite{Scikit2017} to perform a gradient-boosted-tree regression tree search on a hyperparameter space defined in \autoref{tab:hp_space}. In total, we trained 100 models for 30 epochs. \autoref{tab:best_model} yields the set of best hyperparameters found with this method. This model has been further trained to 80 epochs to improve its final accuracy. 

\begin{table}
\begin{minipage}{0.42\textwidth}
	\centering
	\caption{Hyperparameters Space}
	\label{tab:hp_space}
	\begin{tabular}{lll}
		\toprule 
		Parameter & Possible Values \\
		\midrule
		Number of Units		& 128, 256, 512 or 1024 \\
		Number of Layers	& 2, 4 or 6	\\
		Type of Encoder		& LSTM, BLSTM & \\
		Output Dropout		& 0 - 0.9 \\
		State Dropout		& 0 - 0.9 \\
		Variational Dropout  \cite{Gal2016}  & True, False \\
		Learning Rate		& 0.1 - 5 \\
		Decay Factor		& 0.85 - 0.99 \\
		Type of Attention	& ``Luong'' or \\
							& ``Badhanau'' \\
		\bottomrule
	\end{tabular}
\end{minipage}
\hfill
\begin{minipage}{0.43\textwidth}
	\centering
	\caption{Hyperparameter for the best model}
	\label{tab:best_model}
	\begin{tabular}{lll}
		\toprule
		& \textbf{Encoder} & \textbf{Decoder} \\ 
		\midrule
		Number of Units		&	1024	& 2048 \\
		Number of Layers	&	2 & 2 \\
		RNN Cell Type		&	BLSTM	& LSTM \\
		Output Dropout		&	\multicolumn{2}{c}{0.7676} \\	
		State Dropout		&	\multicolumn{2}{c}{0.5374} \\
		Variational Dropout &	\multicolumn{2}{c}{True} \\
		\midrule
		Learning Rate		& \multicolumn{2}{c}{0.355} \\
		Decay Factor		& \multicolumn{2}{c}{0.854} \\
		\bottomrule
	\end{tabular}
	\vspace{5.6mm}
\end{minipage}
\end{table}

\section{Experiments}
\subsection{Reaction prediction}
We evaluated our model on two data sets and compared the performance with other state-of-the-art results. After the hyperparameter optimization, we continued to train our best model on the 395,496 reactions in Jin's USPTO train set and tested the fully trained model on Jin's USPTO test set. Additionally, we trained a second model with the same hyperparameters on 902,581 randomly chosen single-product reactions from the more complex and noisy Lowe dataset and tested it on a set of 50,258 reactions. As molecules are discrete data, changing a single character, such as in source code or arithmetic expressions, can lead to completely different meanings or even invalidate the entire string. Therefore we use full-sequence accuracy, the strictest criteria possible, as our evaluation metric by which a test prediction is considered correct only if all tokens are identical to the ground truth. 

The network has to solve three major challenges. First, it has to memorize the SMILES grammar to predict synthetically correct sequences. Second, because we trained it on canonicalized molecules, the network has to learn the canonical representation. Third, the network has to map the reactants plus reagents space to the product space.
Although the training was performed without a beam search, we used a beam width of 10 without length penalty for the inference. Therefore the 10 most probable sequences were kept at every time step. This allowed us to know what probability the network assigned to each of the sequences. We used the top-1 probabilities to analyze the prediction confidence of the network.

The final step was to canonicalize the network output. This simple and deterministic reordering of the tokens improved the accuracy by 1.5\%. Thus, molecules that were correctly predicted, but whose tokens were not enumerated in the canonical order, were still counted as correct. The prediction accuracies of our model on different datasets are reported in Table \ref{tab:results}. For single product reactions, we achieved an accuracy of 83.2\% on Jin's USPTO test dataset and 65.4\% on Lowe's test set. 


\begin{table}[ht!]
\centering
\caption{Scores of our model on different single product datasets. }
    \begin{tabular}{l r |r r  | r rr }
        \toprule 
        Dataset & Size &  \multicolumn{2}{l}{ } & \multicolumn{3}{l} {Accuracies in [\%]} \\
        && BLEU \cite{Papineni2001} & ROUGE  \cite{Lin}  &  top-1 & top-2 & top-3 \\
        \midrule
        Jin's USPTO test set  \cite{Jin2017} & 38,648 & 95.9 & 96.0& \textbf{83.2} & 87.7 & 89.2 \\
	Lowe's test set \cite{Lowe2017} & 50,258&90.3&90.9&  \textbf{65.4}  & 71.8 & 74.1  \\ 
         \bottomrule
    \end{tabular}
    \label{tab:results}
\end{table}

\subsection{Comparison with the state of the art}

To the best of our knowledge, no previous work has attempted to predict reactions on the complete US patent dataset of Lowe \cite{Lowe2017}.  
Table \ref{tab:comparison} shows a comparison with the Weisfeiler--Lehman difference networks (WLDN) from Jin et al. \cite{Jin2017} on their USPTO test set. To make a fair comparison, we count all the multiple product reactions in the test set as false predictions for our model because we trained only on the single product reactions. By achieving 80.3\% top-1 accuracy, we outperform their model by a margin of 6.3\%, which is even higher than for their augmented setup. As our model does not rank candidates, but was trained on accurately predicting the top-1 outcome, it is not surprising that the WLDN beats our model in the top-3 and top-5 accuracy. The decoding of the 38,648 USPTO test set reactions takes on average 25 ms per reaction, inferred with a beam search. Our model can therefore compete with the state of the art.
\begin{table}[ht!]
\centering
\caption{Comparison with Jin et al. \cite{Jin2017}. The 1,352 multiple product reactions (3.4\% of the test set) are counted as false predictions for our model.}
    \begin{tabular}{l r r r r}
        \toprule 
                   \multicolumn{5}{c}{Jin's USPTO test set \cite{Jin2017}, accuracies in [\%]  }\\
        \midrule
                Method &top-1 & top-2 &   top-3   & top-5  \\
        \midrule
        WLDN  \cite{Jin2017} & 74.0 & & \textbf{86.7} & \textbf{89.5} \\
	Our model & \textbf{80.3} & \textbf{84.7} & 86.2 & 87.5 \\
         \bottomrule
    \end{tabular}
    \label{tab:comparison}
\end{table}
\subsection{Prediction confidence}

\begin{figure}[ht!]
    \begin{subfigure}[b]{0.4\textwidth}
        \includegraphics[width=\textwidth]{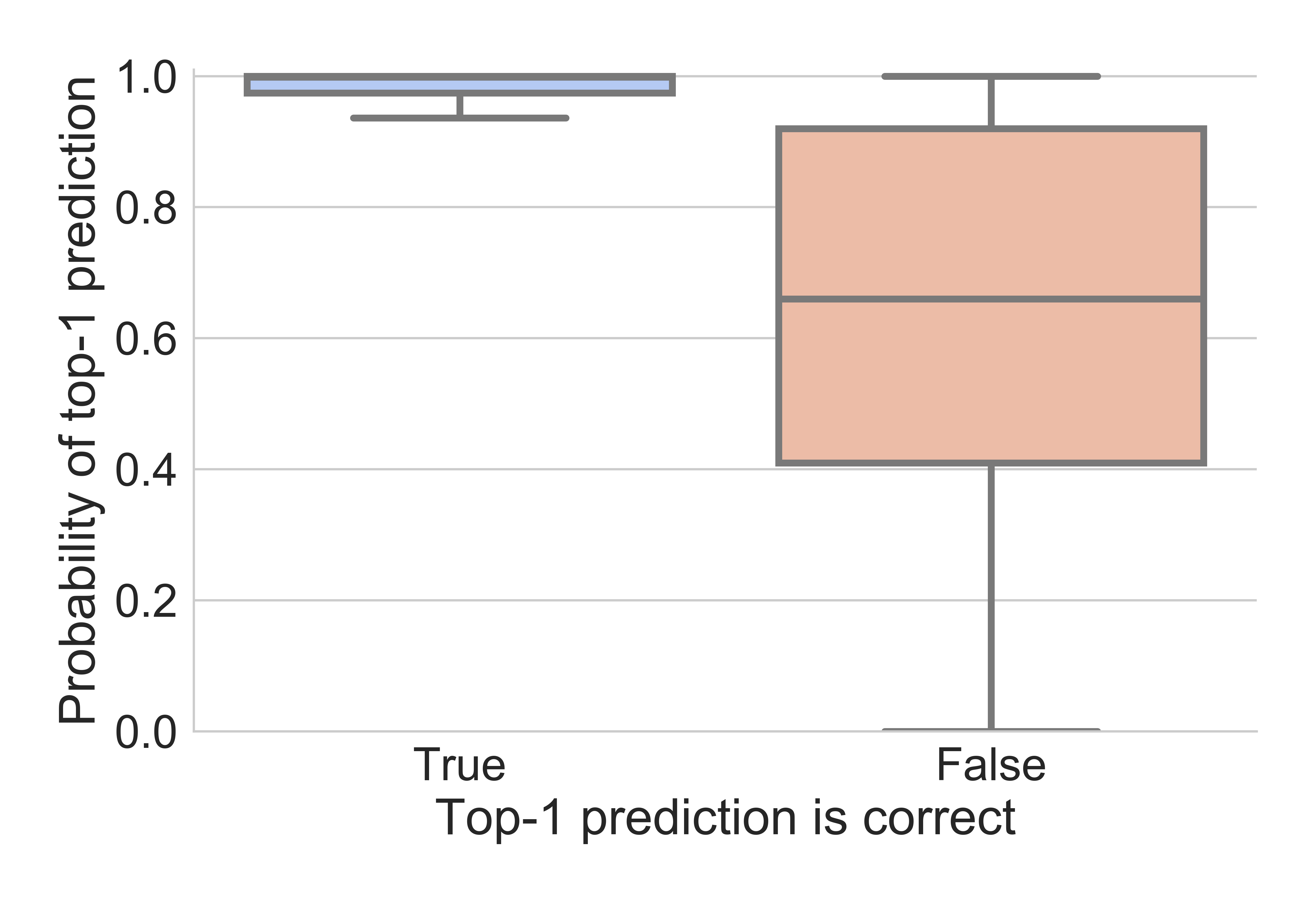}
        \caption{Distribution of top1 probabilities}
        \label{fig:probdistribution}
    \end{subfigure}
    ~ 
    \begin{subfigure}[b]{0.4\textwidth}
        \includegraphics[width=\textwidth]{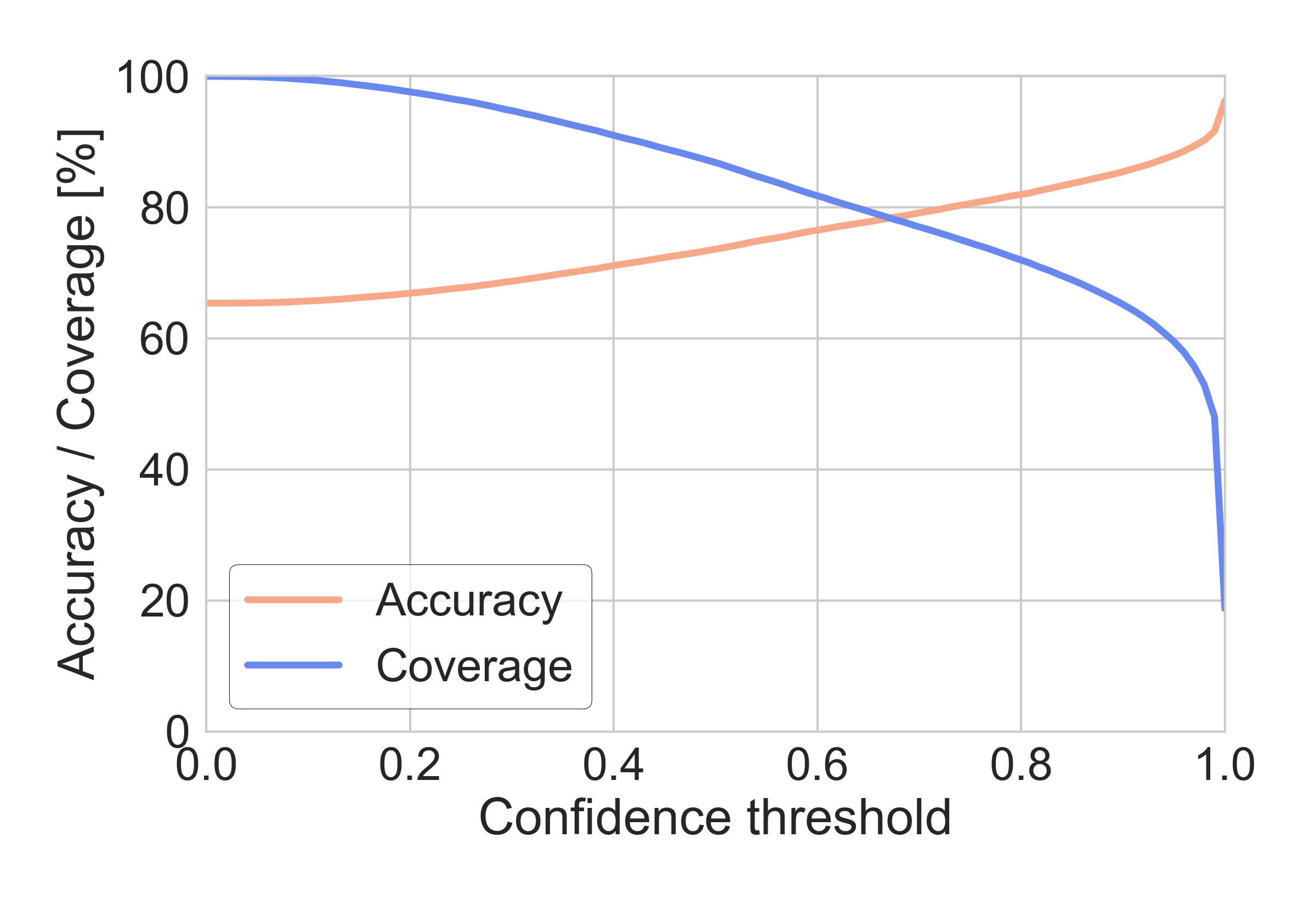}
        \caption{Coverage / Accuracy plot}
        \label{fig:covacc}
    \end{subfigure}
\centering
\caption{Top-1 prediction confidence plots for Lowe's test set inferred with a beam search of 10.}
  \label{fig:top1proba}
\end{figure}

We analyzed the top-1 beam search probability to obtain information about prediction confidence and to observe how this probability was related to accuracy. Figure \ref{fig:probdistribution} illustrates the distribution of the top-1 probability for Lowe's test set in cases where the top-1 prediction is correct (left) and where it is wrong (right). A clear difference can be observed and used to define a threshold under which we determine that the network does not know what to predict. Figure \ref{fig:covacc} shows the top-1 accuracy and coverage depending on the confidence threshold. For example, for a confidence threshold of 0.83 the model would predict the outcome of 70.2\% of the reactions with an accuracy of 83.0\% and for the remaining 29.8\% of the reaction it would not know the outcome. 

\subsection{Attention} 
\begin{figure}[ht!]
    \centering
    \begin{subfigure}[b]{0.4\textwidth}
           \includegraphics[width=\textwidth]{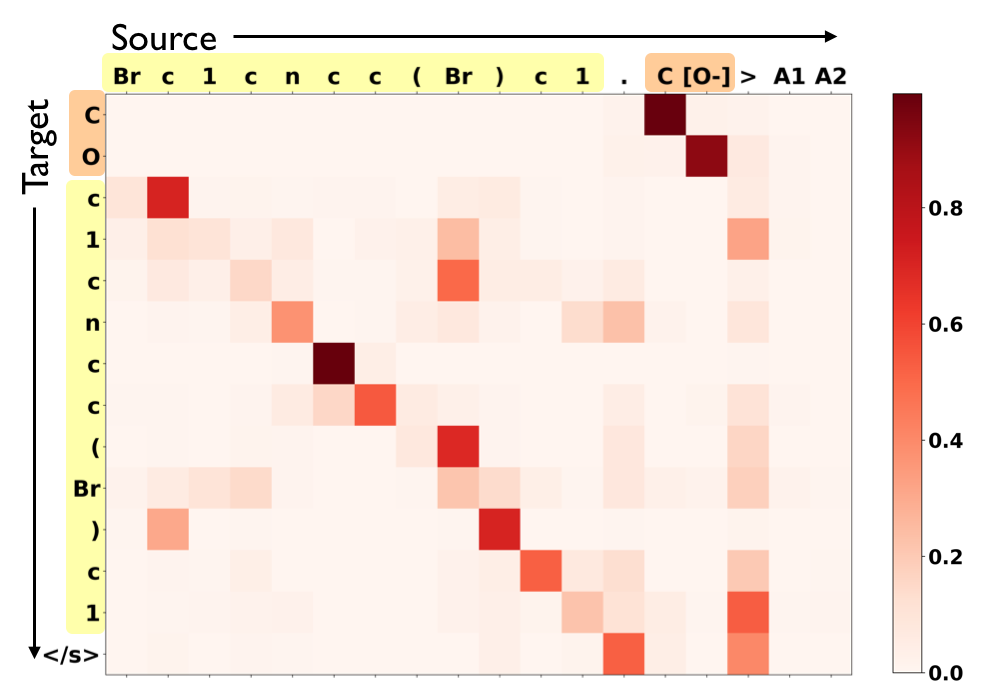}
        \caption{Attention weights}
        \label{fig:alpha}
    \end{subfigure}
    \qquad  \qquad
    \begin{subfigure}[b]{0.4\textwidth}
         \includegraphics[width=\textwidth]{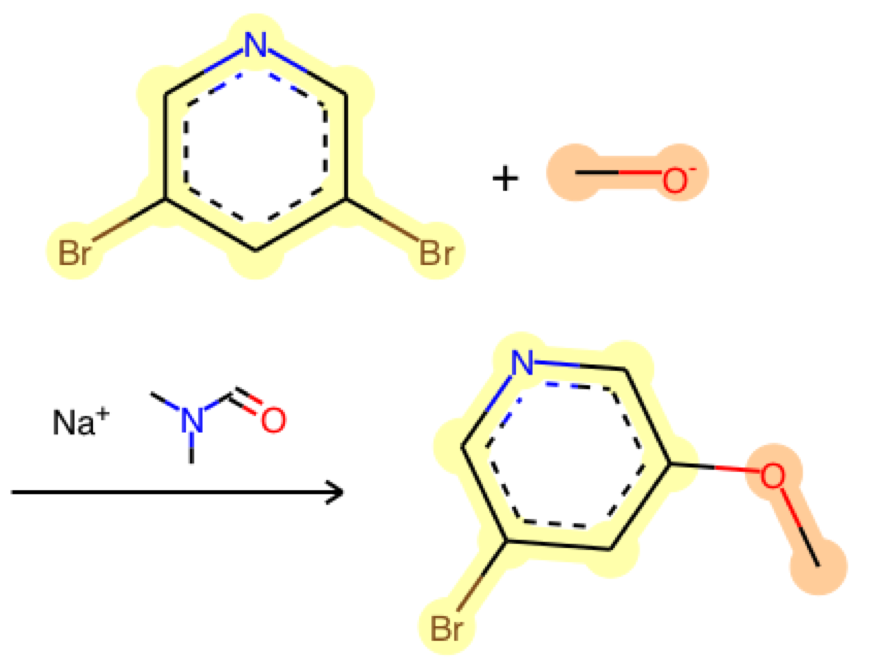}
        \caption{Reaction plotted with RDKit \cite{Landrum2017}}
        \label{fig:lee} 
    \end{subfigure}
    ~ 
    \caption{Reaction 120 from Jin's USPTO test set. The atom mapping between reactants and product is highlighted. SMILES: Brc1cncc(Br)c1.C[O-]$>$CN(C)C=O.[Na+]$>$COc1cncc(Br)c1}
    \label{fig:attention}
\end{figure}

Attention is the key to take into account complex long-range dependencies between multiple tokens. Specific functional groups, solvents or catalysts have an impact on the outcome of a reaction, even if they are far from the reaction center in the molecular graph and therefore also in the SMILES string. Figure \ref{fig:attention} shows how the network learned to focus first on the C[O$^-$] molecule, to map the [O$^-$] in the input correctly to the O in the target, and to ignore the Br, which is replaced in the target.

%
%
\subsection{Limitations}
Our model is not without limitations. An obvious disadvantage compared to template-based methods is that the strings are not guaranteed to be a valid SMILES. Incorporating a context-free grammar layer,  as was done in  \cite{Gomez-Bombarelli2016}, could bring minor improvements. Fortunately, only 1.3\% of the top-1 predictions are grammatically erroneous for our model.  

Another limitation of the training procedure are multiple product reactions. In contrast to words in a sentence, the exact order in which the molecules in the target string are enumerated does not matter. A viable option would be to include in the training set all possible permutations of the product molecules. 

Our hyperparameter space during optimization was restricted to a maximum of 1,024 units for the encoder. Using more units could have led to improvements. On Jin's USPTO dataset, the training plateaued because an accuracy of 99.9\% was reached and the network had memorized almost the entire training set. Even on Lowe's noisier dataset, a training accuracy of 94.5\% was observed. A hyperparameter optimization should be performed on Lowe's dataset. 



\section{Conclusion}
Predicting reaction outcomes is a routine task of many organic chemists trained to recognize structural and reactivity patterns reported in a wide number of publications. Not only did we show that a seq2seq model with correctly tuned hyperparameters can learn the language of organic chemistry, our approach also outperformed the current state-of-the-art in patent reaction outcome prediction by achieving 80.3\% on Jin's USPTO dataset and 65.4\% on single product reactions of Lowe's dataset. Compared to previous work, our approach is fully data driven and free of reaction templates. Also worth mentioning is the overall simplicity of our model that jointly trains the encoder, decoder and attention layers. Our hope is that, with this type of model, chemists can codify and perhaps one day fully automate the art of organic synthesis.




\subsubsection*{Acknowledgments}
We thank Nadine Schneider, Greg Landrum and Roger Sayle for the helpful discussions on RDKit and the datasets. We also would like to acknowledge Marwin Segler and Hiroko Satoh for useful feedback on our approach.

\small
\bibliographystyle{naturemag}
\bibliography{biblio,biblio_arxiv,biblio_suppl}

\end{document}